\documentclass{article}

\usepackage[dblblindworkshop,final]{neurips_2026}
\workshoptitle{Can We Trust the Judge? (JUDGe) @ NeurIPS 2026}

\usepackage[utf8]{inputenc}
\usepackage[T1]{fontenc}
\usepackage{hyperref}
\usepackage{url}
\usepackage{booktabs}
\usepackage{amsfonts}
\usepackage{amsmath}
\usepackage{nicefrac}
\usepackage{microtype}
\usepackage{xcolor}
\usepackage{graphicx}

\definecolor{tolBlue}{HTML}{0077BB}    
\definecolor{tolOrange}{HTML}{EE7733}  
\definecolor{tolGrey}{HTML}{BBBBBB}    

\title{The First Token Is Not the Verdict:\\
Hidden Costs of Reading LLM Judges Without Generating}

\author{%
  Gnaneswar Villuri \quad Hashmath Shaik \quad Alex Doboli \\
  Department of Electrical and Computer Engineering \\
  Stony Brook University \\
  Stony Brook, NY 11794, USA
}

\begin{document}
\maketitle

\begin{abstract}
Reading an LLM judge's verdict from the logits of its first generated token is
cheap, requires no generation, and is exactly what constrained decoding and
likelihood-scoring evaluation harnesses produce. We show that this readout
distorts position bias in one direction: it overstates it in every condition we
test, so figures obtained this way behave as upper bounds. The mechanism is that
judges do not always lead with a verdict token, on $12\%$ to $49\%$ of pairs for
three Qwen3 judges and under $3\%$ for Llama-3.1-8B and Phi-3.5-mini, and forcing a
read on those pairs returns whichever response was shown first rather than a
judgment. Pooled over the $924$ pairs where a judge did not commit, the forced read
flips on $89.7\%$ of them when the responses are swapped, against $47.5\%$ read
after generation (paired difference $+0.422$, $95\%$ CI $[+0.365, +0.467]$). The
distortion is specific to what is measured: it moves position bias by $42$ points
while moving judge accuracy by under one point in seven of ten conditions, so it
misleads whoever audits a judge rather than whoever uses one. A second, smaller
failure occurs even when the judge does lead with a verdict token, since it
sometimes opens with one letter and reasons its way to the other, on $0$ to $5.5\%$
of pairs at a rate uncorrelated with compliance. We recommend reporting the rate at
which a judge leads with a verdict token, which costs one forward pass and no
labels, alongside any position-bias figure.
\end{abstract}

\section{Two protocols, one name}

Position bias is measured by presenting each pair of responses in both orders and
counting how often the verdict \emph{flips}, that is, reverses
\citep{zheng2023judging,norman2026reliability}.
Obtaining that verdict requires choosing a protocol, and the two protocols in
common use are not the same operation.

The first protocol generates a response and parses the verdict from the text,
which is what judge audits do \citep{norman2026reliability,darkcurrent2026,tan2024judgebench}.
The second reads the verdict from the logits of the first generated token. It is
far cheaper because it requires no generation at all, it is what
restricted-likelihood evaluation harnesses compute, and it is what constrained
decoding produces by construction, since token masking guarantees a legal verdict
at the first position \citep{eval-harness,willard2023outlines}. That guarantee is
advertised as the point: one never gets a parse error and never has to retry.

Existing practice handles a non-verdict first token in one of three ways, and two
of them are the forced read. Restricted-likelihood harnesses score only the
candidate answers against each other and never ask what the model would have said
\citep{eval-harness}. Constrained decoding masks every illegal token, so a verdict
is produced by construction \citep{willard2023outlines}. The third way is to
generate and, on a parse failure, retry or let the judge continue, which is what
JudgeBench does \citep{tan2024judgebench}. Only the third recovers what the judge
would have answered.

Logit-based judging is in production use rather than hypothetical.
Probability-weighted score aggregation is standard \citep{liu2023geval}, and
AlpacaEval~2.0 computes its headline win-rate from the logprobs of a single
output token \citep{alpacaeval}. We did not identify a published position-bias
figure obtained this way, so we ask a conditional question: if judge
verdicts are obtained as this tooling obtains them, what does that cost when the
quantity measured is position bias? The guarantee turns out to be the problem.
A judge that was not going to commit still yields a verdict, and that verdict
carries position rather than content.

\section{Setup}

A \emph{pair} is two responses to one instruction, shown in two
\emph{presentations}, one per order, under a prompt ending ``Answer with exactly
one character''. We take two readouts of the same judge on the same pairs. The
\textbf{first-token} readout takes the larger of the two verdict-token logits at
the first generated position, whether or not either was what the judge would
actually have emitted; we call it the \emph{forced read}, and that property is
what this paper is about. The \textbf{generated} readout lets the model write up
to $500$ tokens and parses the verdict using explicit decision patterns such as
``the answer is X'', rather than bare mentions of the letters, because the judge
names both responses repeatedly while reasoning; where none parses we regenerate
once at $1{,}500$ tokens. A pair is \emph{compliant} when
the judge would have led with a verdict token in both presentations, that is, when
the argmax over its full vocabulary is one of the two verdict tokens.

The judges are Qwen3 at $1.7$B, $4$B and $8$B, Llama-3.1-8B-Instruct and
Phi-3.5-mini. The benchmarks are JudgeBench \citep{tan2024judgebench}, $620$
pairs, and LLMBar \citep{zeng2024llmbar}, $419$ pairs. Verdict tokens are X/Y, A/B
and 1/2. We run the full benchmarks. Intervals are cluster bootstraps over
benchmark sources. Qwen3 supports a hybrid thinking mode, which we
disable (\texttt{enable\_thinking=False}) so that all judges receive comparable
single-turn prompts; the non-compliance we report is therefore not an artefact of
an explicit reasoning mode being left on.

Two properties of this design make our numbers conservative. Our prompt asks for
one character, so the generated readout measures the judge disobeying it; audits
that invite reasoning would see lower compliance and a larger divergence. And the $88$ presentations that still yield no verdict at $1{,}500$ tokens are
excluded from both readouts; every such pair is non-compliant, so the exclusion
removes data from the arm that produces the effect. Recovery is uneven: the second
pass resolves $95\%$ of the Qwen3-8B cases but $54\%$ for Qwen3-1.7B, which often
reaches the cap without concluding.

\section{Compliance varies widely and is not predictable}

\begin{table}[t]
  \caption{Compliance is the fraction of pairs on which the judge would have led
  with a verdict token in both orders; Gap is the flip-rate difference between the
  forced first-token readout and the selective readout of Section~\ref{sec:cheap}.
  \textbf{The selective column is not a generate-everywhere baseline}, so the
  $+0.000$ rows are definitional rather than evidential; Section~\ref{sec:threepc}
  measures how often the substitution is wrong. Compliance is over all pairs run,
  flip rates and $n$ over pairs where both readouts yield a verdict, and every
  excluded pair is non-compliant. Two further conditions with gap $0.000$
  are omitted for space.}
  \label{tab:main}
  \centering
  \small
  \begin{tabular}{llccccl@{\hspace{4pt}}r}
    \toprule
    Judge & Bench & Labels & Compl. & First & Selective & Gap [95\% CI] & $n$ \\
    \midrule
    Qwen3-8B     & LLMBar     & X/Y & $1.000$ & $0.320$ & $0.320$ & $+0.000$ [$0.000$, $0.000$] & $419$ \\
    Llama-3.1-8B & JudgeBench & X/Y & $0.998$ & $0.582$ & $0.582$ & $+0.000$ [$0.000$, $0.000$] & $619$ \\
    Phi-3.5-mini & JudgeBench & X/Y & $0.971$ & $0.544$ & $0.542$ & $+0.002$ [$0.000$, $+0.005$] & $616$ \\
    Qwen3-4B     & JudgeBench & X/Y & $0.885$ & $0.658$ & $0.594$ & $+0.065$ [$+0.019$, $+0.102$] & $620$ \\
    Qwen3-8B     & JudgeBench & 1/2 & $0.676$ & $0.905$ & $0.783$ & $+0.123$ [$+0.037$, $+0.181$] & $612$ \\
    Qwen3-1.7B   & JudgeBench & X/Y & $0.663$ & $0.875$ & $0.787$ & $+0.088$ [$+0.041$, $+0.134$] & $567$ \\
    Qwen3-8B     & JudgeBench & A/B & $0.661$ & $0.847$ & $0.700$ & $+0.147$ [$+0.068$, $+0.211$] & $614$ \\
    Qwen3-8B     & JudgeBench & X/Y & $0.506$ & $0.731$ & $0.511$ & $+0.220$ [$+0.118$, $+0.302$] & $614$ \\
    \bottomrule
  \end{tabular}
\end{table}

Compliance is a property of a model, a prompt and a task together, not of a model
alone. Holding Qwen3-8B and the benchmark fixed and changing only the two verdict
tokens moves it from $0.506$ to $0.676$. Holding the model and the prompt fixed
and changing the benchmark moves it from $0.506$ to $1.000$, LLMBar items being
short instruction-following comparisons while JudgeBench items are technical
questions that invite step-by-step work. Across Qwen3 sizes compliance is
non-monotone, running $0.663$, $0.885$, $0.506$ at $1.7$B, $4$B and $8$B. Reported elsewhere for this same Llama-3.1-8B under a
multiple-choice prompt, first-token validity is $9.5\%$ \citep{prefilling2025},
against $99.8\%$ under ours. A compliance rate cannot be inherited from another
study.

\section{A forced read on an undecided judge locks to a slot}

\begin{figure}[t]
\centering
\includegraphics[width=\linewidth]{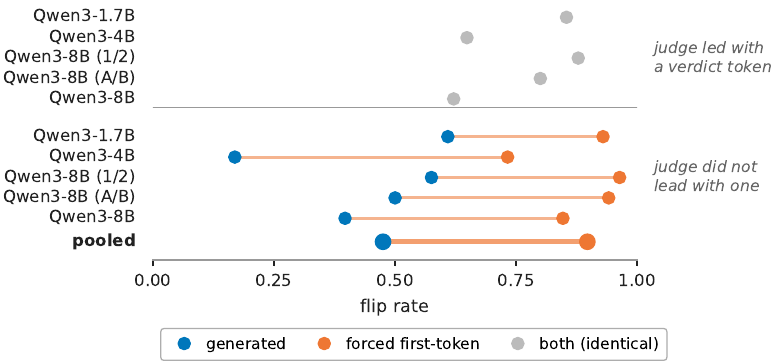}
\caption{Where the two readouts differ, for the five Qwen3--JudgeBench conditions
of Table~\ref{tab:main}. Where the judge leads with a verdict token (top) they are
identical \emph{by construction}, so one marker is shown. Where it does not
(bottom) the forced read always lies to the right of the generated read. Pooled
over $924$ such pairs the difference is $+0.422$ ($p = 1.0\times10^{-99}$).}
\label{fig:dumbbell}
\end{figure}

The two readouts diverge only on pairs where the judge did not lead with a
verdict token, and there they diverge sharply. Because both readouts are measured
on the same pairs, the comparison is paired, and we pool it across conditions. Of
$924$ pooled non-compliant pairs, the forced read flips on $829$ of them
($0.897$) and the generated read on $439$ ($0.475$), a paired difference of
$+0.422$ with a $95\%$ cluster-bootstrap interval of $[+0.365, +0.467]$ over
conditions. Of the $420$ pairs on which the two readouts disagree about whether a
flip occurred, $405$ are flips seen only under the forced read and $15$ only under
generation (exact McNemar $p = 1.0\times10^{-99}$, which treats pairs as
independent and so understates the true $p$; the clustered interval above is the
one we rely on).

What the forced read is measuring follows by identity rather than by inference.
With two verdict tokens and no ties, a flip occurs exactly when the same slot
position is selected in both presentations, because the responses swap between
them; we verify that this holds on all $5{,}720$ pairs (the $11{,}596$ order-presentations
counted in Section~\ref{sec:cheap} are the $5{,}798$ pairs before exclusion, each
presented twice). The flip rate on
non-compliant pairs is therefore not evidence of a slot lock but is itself the
rate at which the forced read locked to a slot, $0.897$.

The slot it locks to is almost always the first one. Of the $829$ locked
non-compliant pairs, $827$ locked to the first position ($0.998$; exact binomial
against chance $p = 1.9\times10^{-244}$), and four of the five conditions lock to
it without a single exception. The lock is to a fixed position at the
logit level rather than being item-specific, and it makes the error one-sided:
in all five conditions the forced read flips more, never less, so position-bias
figures obtained from first-token logits are upper bounds on what a
generate-and-parse audit would report.

The choice of readout affects measurement rather than use. Against gold labels
the two readouts differ in accuracy by nothing at all in four conditions and by
more than a point in only three, the largest difference being $6.7$ points
($0.439$ to $0.648$ overall). Flip rate, by contrast, differs by up to $0.220$. Both readouts are
near chance on the non-compliant pairs that drive the divergence, because the
judge does not know the answer there under either protocol. The forced
read adds error that depends on order but not on the label, so it affects whoever
audits a judge far more than whoever uses one.

\paragraph{A compliant first token is usually, but not always, the conclusion.}
\label{sec:threepc} On
$300$ compliant pairs per judge we compared the first token to the parsed final
verdict. Qwen3-4B disagrees on $15$ of $271$ parsed ($5.5\%$), Phi-3.5-mini on
$14$ of $257$ ($5.4\%$), Qwen3-8B on $8$ of $285$ ($2.8\%$) and Llama-3.1-8B on
$0$ of $300$ ($0.0\%$, Wilson $95\%$ CI $[0.0\%, 1.3\%]$). The disagreements share
a form: the judge opens with \textbf{X}, then writes ``\ldots Therefore,
\textbf{Response Y} better follows the instruction.'' This failure is independent
of the first: across the four judges its rate is uncorrelated with compliance
($r = 0.01$), and Phi-3.5-mini leads with a verdict token on $97.1\%$ of pairs yet
changes its mind on $5.4\%$ of them. The judge that looks safest for a first-token
readout is not the one least distorted by it.

\section{A cheaper procedure, and what it is worth}
\label{sec:cheap}

The characterisation suggests a shortcut: read the first token, and generate only
for the orders where the judge did not lead with a verdict. Adaptive computation of
this shape is standard \citep{jung2024trustescalate,gupta2024cascades}, but those
cascades escalate to a stronger model under a calibrated threshold, whereas we
escalate between two readouts of the same model and need no calibration set.
The saving is large and the fidelity is high but not perfect. Across our ten
conditions the procedure generates for $1{,}678$ of $11{,}596$ order-presentations,
a reduction of $85.5\%$, ranging from $58.9\%$ on the least compliant condition to
$100\%$ where the judge always leads with a verdict. By Section~\ref{sec:threepc} the substituted verdicts are wrong on $0$ to $5.5\%$
of compliant pairs, so fidelity runs from about $0.945$ to $1.00$. It is an
approximation with a measured error rate rather than a guarantee.

\section{Limitations}

The divergence is measurable in only one of the three families we test. Compliance
is so high for the other two that Phi-3.5-mini contributes $18$ non-compliant
pairs and Llama-3.1-8B one, too few to analyse, so the pooled comparison is Qwen3
only and the cross-family contrast is a difference in compliance rather than an
independent replication of the effect. We also test only open-weight judges at
$8$B and below, because frontier API judges do not expose logits.

Two measurement choices should be read with the result. The generated readout uses
a heuristic parser and a capped budget, so even after the second pass it is a
lower bound on how often a verdict is eventually reached. And we tested whether the two readouts reorder judges by position bias; the
reordering was not stable under bootstrap ($50.1\%$ of resamples) and we do not
claim it.

Our question is conditional on a practice rather than a correction of a published
number. That first-token reads fail when a model does not commit is already stated
for multiple-choice settings \citep{prefilling2025}, and the readout is in
production use \citep{alpacaeval}. Prompts engineered for single-token output
may reach compliance close to $1$, and output prefilling \citep{prefilling2025} is
a plausible remedy we have not tested; our point is that neither is usually
verified or reported.

\section{Recommendations and conclusion}

Three things belong in a judge deployment disclosure, all cheap and currently
absent from reported evaluations. Report the compliance rate: it costs one forward
pass per pair, needs no labels, varies from $0.506$ to $1.000$ across our
conditions, and cannot be inherited from another paper.
\citet{agreement2026} ask the same of abstention and tie rates in generated
outputs; ours applies that discipline one step earlier. Say which readout produced the verdict, because
a figure obtained by constrained decoding and one obtained by generate-and-parse
are not the same quantity when compliance is incomplete. And treat the answer
format as an experimental variable: changing two characters moves compliance by
$17$ points here, and most papers do not report which verdict tokens they used.

The first token is a convenient place to read a judge's verdict and an unreliable
one. It is not a verdict at all on up to $49\%$ of pairs, and not its conclusion on
up to $5.5\%$ of those where it looks like one. Both rates are measurable in
advance and neither is reported.

\bibliographystyle{plainnat}
\bibliography{references}

@inproceedings{zheng2023judging,
  title     = {Judging {LLM}-as-a-Judge with {MT-Bench} and {Chatbot Arena}},
  author    = {Zheng, Lianmin and Chiang, Wei-Lin and Sheng, Ying and Zhuang, Siyuan
               and Wu, Zhanghao and Zhuang, Yonghao and Lin, Zi and Li, Zhuohan
               and Li, Dacheng and Xing, Eric P. and Zhang, Hao and
               Gonzalez, Joseph E. and Stoica, Ion},
  booktitle = {Advances in Neural Information Processing Systems (Datasets and Benchmarks)},
  year      = {2023}
}

@inproceedings{tan2024judgebench,
  title     = {{JudgeBench}: A Benchmark for Evaluating {LLM}-based Judges},
  author    = {Tan, Sijun and Zhuang, Siyuan and Montgomery, Kyle and
               Tang, William Y. and Cuadron, Alejandro and Wang, Chenguang and
               Popa, Raluca Ada and Stoica, Ion},
  journal   = {arXiv preprint arXiv:2410.12784},
  year      = {2024}
}

@inproceedings{zeng2024llmbar,
  title     = {Evaluating Large Language Models at Evaluating Instruction Following},
  author    = {Zeng, Zhiyuan and Yu, Jiatong and Gao, Tianyu and Meng, Yu and
               Goyal, Tanya and Chen, Danqi},
  booktitle = {International Conference on Learning Representations (ICLR)},
  year      = {2024}
}

@inproceedings{jung2024trustescalate,
  title     = {Trust or Escalate: {LLM} Judges with Provable Guarantees for
               Human Agreement},
  author    = {Jung, Jaehun and Brahman, Faeze and Choi, Yejin},
  booktitle = {International Conference on Learning Representations (ICLR)},
  year      = {2025},
  note      = {arXiv:2407.18370}
}

@article{gupta2024cascades,
  title   = {Language Model Cascades: Token-level Uncertainty and Beyond},
  author  = {Gupta, Neha and Narasimhan, Harikrishna and Jitkrittum, Wittawat
             and Rawat, Ankit Singh and Menon, Aditya Krishna and Kumar, Sanjiv},
  journal = {arXiv preprint arXiv:2404.10136},
  year    = {2024}
}

@article{prefilling2025,
  title   = {Improving {LLM} First-Token Predictions in Multiple-Choice Question
             Answering via Output Prefilling},
  author  = {Cappelletti, Silvia and Poppi, Tobia and Poppi, Samuele and
             Yong, Zheng-Xin and Garcia-Olano, Diego and Cornia, Marcella and
             Baraldi, Lorenzo and Cucchiara, Rita},
  journal = {arXiv preprint arXiv:2505.15323},
  year    = {2025}
}

@article{norman2026reliability,
  title   = {Reliability without Validity: A Systematic, Large-Scale Evaluation
             of {LLM}-as-a-Judge Models Across Agreement, Consistency, and Bias},
  author  = {Norman, Justin D. and Rivera, Michael U. and Hughes, D. Alex},
  journal = {arXiv preprint arXiv:2606.19544},
  year    = {2026}
}

@article{darkcurrent2026,
  title   = {{LLM} Judges Have Dark Current: A Psychometric Datasheet for
             {LLM}-as-a-Judge Evaluation},
  author  = {Usami, Hiroyasu and Hara, Keisuke and Tsuboi, Ayato and
             Matsuda, Naohiko},
  journal = {arXiv preprint arXiv:2606.15610},
  year    = {2026}
}

@misc{eval-harness,
  title  = {A Framework for Few-shot Language Model Evaluation},
  author = {Gao, Leo and Tow, Jonathan and Abbasi, Baber and Biderman, Stella
            and others},
  year   = {2024},
  howpublished = {\url{https://github.com/EleutherAI/lm-evaluation-harness}}
}

@article{willard2023outlines,
  title   = {Efficient Guided Generation for Large Language Models},
  author  = {Willard, Brandon T. and Louf, R{\'e}mi},
  journal = {arXiv preprint arXiv:2307.09702},
  year    = {2023}
}

@misc{alpacaeval,
  title  = {{AlpacaEval}: An Automatic Evaluator of Instruction-following Models},
  author = {Li, Xuechen and Zhang, Tianyi and Dubois, Yann and Taori, Rohan and
            Gulrajani, Ishaan and Guestrin, Carlos and Liang, Percy and
            Hashimoto, Tatsunori B.},
  year   = {2023},
  howpublished = {\url{https://github.com/tatsu-lab/alpaca_eval}},
  note   = {AlpacaEval 2.0 weighted win-rate uses single-token logprobs}
}

@article{agreement2026,
  title   = {Agreement Metrics for {LLM}-as-Judge Evaluation: What to Report
             and Why},
  author  = {Rao, Delip and Callison-Burch, Chris},
  journal = {arXiv preprint arXiv:2606.00093},
  year    = {2026}
}

@inproceedings{liu2023geval,
  title     = {{G-Eval}: {NLG} Evaluation using {GPT-4} with Better Human Alignment},
  author    = {Liu, Yang and Iter, Dan and Xu, Yichong and Wang, Shuohang and
               Xu, Ruochen and Zhu, Chenguang},
  booktitle = {Proceedings of EMNLP},
  year      = {2023}
}

\newpage

\end{document}